\documentclass[letterpaper, 10 pt, conference]{ieeeconf}  

\IEEEoverridecommandlockouts                              

\usepackage{graphics}
\usepackage[pdftex]{graphicx}
\usepackage[tight,footnotesize]{subfigure}
\usepackage{multirow}
\usepackage[cmex10]{amsmath}
\usepackage{amssymb}
\usepackage{amsfonts}
\usepackage{algpseudocode}
\usepackage{algorithm}
\usepackage{booktabs}
\usepackage{url}
\usepackage{color}
\usepackage{hyperref}

\title{\LARGE \bf
SoMaSLAM: 2D Graph SLAM for Sparse Range Sensing\\
with Soft Manhattan World Constraints
}

\author{Jeahn Han$^{1}$, Zichao Hu$^{2}$, Seonmo Yang$^{1}$, Minji Kim$^{1}$, and Pyojin Kim$^{1}$
\thanks{$^{1}$School of Mechanical and Robotics Engineering, Gwangju Institute of Science and Technology (GIST), Gwangju 61005, South Korea. {\tt\small \{jeahnhaan, seonmo.yang, minji0110\}@gm.gist.ac.kr}, {\tt\small \{pjinkim\}@gist.ac.kr}}%
\thanks{$^{2}$Department of Computer Science, University of Texas at Austin, Texas 78712, United States of America. {\tt\small \{zichao\}@utexas.edu}}%
}

\begin{document}

\maketitle
\thispagestyle{empty}
\pagestyle{empty}

\begin{abstract}
We propose a graph SLAM algorithm for sparse range sensing that incorporates a soft Manhattan world utilizing landmark-landmark constraints.
Sparse range sensing is necessary for tiny robots that do not have the luxury of using heavy and expensive sensors.
Existing SLAM methods dealing with sparse range sensing lack accuracy and accumulate drift error over time due to limited access to data points.
Algorithms that cover this flaw using structural regularities, such as the Manhattan world (MW), have shortcomings when mapping real-world environments that do not coincide with the rules.
We propose SoMaSLAM, a 2D graph SLAM designed for tiny robots with sparse range sensing.
Our approach effectively maps sparse range data without enforcing \textit{strict} structural regularities and maintains an adaptive graph.
We implement the MW assumption as soft constraints, which we refer to as a \textit{soft} Manhattan world.
We propose novel soft landmark-landmark constraints to incorporate the soft MW into graph SLAM.
Through extensive evaluation, we demonstrate that our proposed SoMaSLAM method improves localization accuracy on diverse datasets and is flexible enough to be used in the real world.
We release our source code and sparse range datasets at {\textcolor{magenta}{\url{https://SoMaSLAM.github.io/}}}.
\end{abstract}

\section{Introduction}
\label{sect:intro}


The range sensor is one of the most effective sensors in simultaneous localization and mapping (SLAM)~\cite{zhang2014loam, shan2018lego} for accurately recognizing the surrounding environments.
Although heavy, bulky, and expensive range sensors, such as Velodyne VLP-16, are very accurate and reliable, they are sometimes not suitable for specific situations.
Palm-sized tiny robots like Crazyflie 2.1~\cite{giernacki2017crazyflie} cannot utilize these range sensors due to their limited payload, supporting a maximum weight of 15 g.
In these cases, it is logical to sacrifice accuracy and the number of range sensing points for a lightweight range sensor, which leads to the sparse range sensing problem.

Sparse range sensing severely limits the range data, meaning existing SLAM methods not tailored to the problem cannot work properly due to the lack of information.
Several prior works~\cite{beevers2006slam, zhou2022efficient} tackle this challenge to overcome limited range sensing capacity within particle filter and pose graph SLAM frameworks.
Despite these efforts, drift error accumulates over time and the methods mentioned do not take advantage of repeated structural regularities.

Some SLAM methods~\cite{zou2019structvio, joo2021linear, jeong2023linear} integrate structural regularities such as Manhattan world (MW)~\cite{coughlan1999manhattan} where all planes must be parallel or orthogonal to each other.
These algorithms neutralize the drift error by effectively utilizing indoor structural patterns and consistently produce accurate results but fail when pre-assumed structural models are broken.
An open door or an out-of-place desk can be enough to disrupt the regulated conditions of a Manhattan world.
Other structural models like Atlanta world~\cite{schindler2004atlanta} or a mixture of Manhattan frames~\cite{Straub_2014_CVPR} allow a bit more flexibility.
Nonetheless, the fact that all features must obey a mandatory rule when using structural regularities makes SLAM algorithms vulnerable to failure when using real-world datasets.

\begin{figure}[t]
\centering
\includegraphics[width=0.95\linewidth]{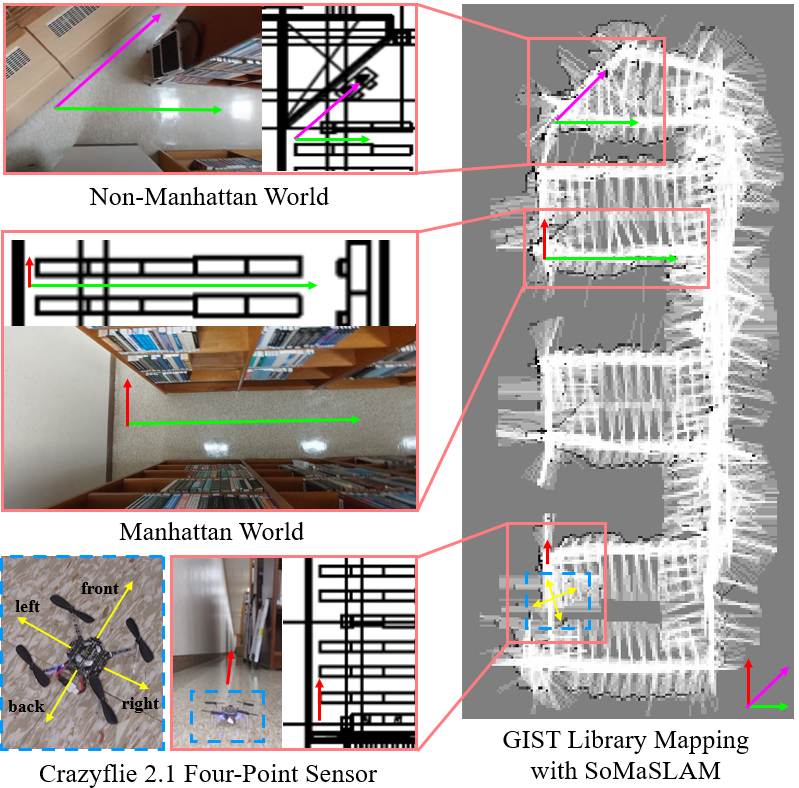}
\vspace{-1.0mm}
\caption{Our SoMaSLAM results (right) where non-Manhattan (top left) and Manhattan (left-middle) worlds co-exist in a structured environment.
We obtain sparse data by a low-cost, lightweight (2.3 g), four radially-spaced ToF range sensor attached to the Crazyflie~\cite{giernacki2017crazyflie} nano drone (bottom left).}
\vspace{-4.0mm}
\label{fig:front_page_image}
\end{figure}

To address these issues, we propose SoMaSLAM, a novel 2D graph SLAM for sparse range sensing leveraging structural regularities, which extends the previous 2D graph SLAM~\cite{zhou2022efficient} to achieve more accurate and consistent performance in structured environments.
In particular, we formulate the objective function in pose graph optimization (PGO) that \textit{encourages}, rather than forces, nearby walls to be orthogonal or parallel to each other.
We propose new soft constraints~\cite{thrun2006graph} between landmarks called the \textit{soft} Manhattan world (MW)~\cite{tsai2011real}, devoid of \textit{hard} constraints~\cite{liu2020visual, jeong2023linear}.
Our soft MW approach in graph SLAM allows more flexibility in structured environments and molds the map into a version that best satisfies all conditions as shown in Fig.~\ref{fig:front_page_image}.
Our main contributions are as follows:
\begin{itemize}
\item[$\bullet$] We leverage structural regularities in man-made environments for a sparse range sensing problem in the pose graph SLAM framework.
\item[$\bullet$] We propose a new soft landmark-landmark constraint on the PGO-based soft MW, allowing more flexibility in structured environments.
\item[$\bullet$] We evaluate the proposed SoMaSLAM on author-collected and public Radish~\cite{Radish} datasets and compare to other state-of-the-art SLAM methods.
We also make author-collected datasets and code publicly available.
\end{itemize}
To the best of our knowledge, this is the first 2D graph SLAM with soft constraints between landmarks inspired by the soft MW assumption, which can be violated but violating the soft constraints incurs a penalty in the objective function in PGO.



\section{Related Work}
\label{sect:relatedwork}


Graph SLAM, which frames SLAM as a graph, is a popular approach for its nonlinear optimization and flexible graph structure.
We dive deeper into SLAM with sparse range sensing, and SLAM with structural regularities such as the Manhattan world assumption~\cite{coughlan1999manhattan}.

SLAM with sparse sensing is a category of SLAM where the robot receives and uses very few data points from sensors, often resulting in challenges related to localization and mapping accuracy.
Despite the need for SLAM with sparse sensing, the field has few published works.
Beevers et al.~\cite{beevers2006slam} addresses the SLAM problem with sparse sensing, utilizing a Rao-Blackwellized particle filter~\cite{murphy2001rao} to combine data from several scans to increase density.
Yap et al.~\cite{yap2009slam} tackle SLAM using low-cost, noisy sonar sensors by employing particle filters and a line-segment-based map with an orthogonality assumption.
Zhou et al.~\cite{zhou2022efficient} addresses SLAM with sparse sensing by utilizing a graph SLAM with a novel front-end to replace scan matching and improved loop closing in the back-end, specifically designed for sparse and uncertain range data.
In NanoSLAM~\cite{niculescu2023nanoslam}, new high-density ToF sensor hardware is introduced for Crazyflie nano drones~\cite{giernacki2017crazyflie} to overcome sparse sensing.
Despite these approaches, SLAM with sparse sensing frequently suffers from a lack of accuracy due to the inherent limitation of relying on minimal data points.

Some odometry and SLAM methods incorporate structural regularities to maximize performance at the cost of being able to map diverse environments.
The Manhattan world~\cite{coughlan1999manhattan} is a fundamental structural regularity and a widely favored choice for integration into SLAM algorithms.
Liu et al.~\cite{liu2020visual} present a stereo visual SLAM for man-made environments, utilizing a two-stage Manhattan frame (MF) tracking with line features for absolute orientation estimation based on ORB-SLAM2~\cite{mur2017orb}.
Jeong et al.~\cite{jeong2023linear} outlines a SLAM algorithm that supplements the limitations of RGB-D cameras using a four-point LiDAR, helping to build a reliable global MW map and estimate 6-DoF camera poses.
A step up from the Manhattan world is the Atlanta world (AW)~\cite{schindler2004atlanta}.
In an Atlanta world, all walls remain orthogonal to the ground plane, but no longer need to maintain a fixed relationship (orthogonality) with one another.
Joo et al.~\cite{joo2021linear} propose a linear RGB-D SLAM designed for the Manhattan and Atlanta world by accommodating a vertical axis and multiple horizontal directions with planar features in the Kalman filter framework.
Yet, these SLAM methods that utilize structural regularities, such as MW and AW, lack flexibility in representing the surrounding structured environments.
The environment must suit the rules perfectly for the methods to perform well.
They are prone to failure when mapping non-structurally regulated environments.
Moreover, much of the discussed methods rely on an RGB-D camera, which is often unavailable for nano drones~\cite{niculescu2023nanoslam} like Crazyflie.

\section{Proposed Method}
\label{sect:proposedmethod}

\begin{figure*}[t]
\centering
\includegraphics[width=\textwidth]{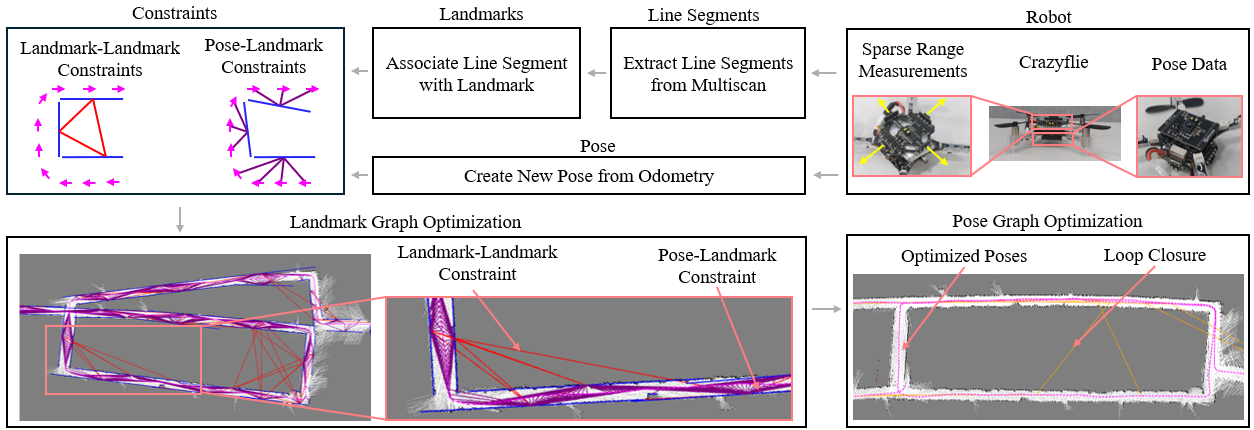}
\vspace{-6mm}
\caption{Overview of the proposed SoMaSLAM.
We construct poses (magenta arrows) and landmarks (blue) in graph SLAM from odometry and line segments given sparse range measurements.
We generate and impose pose-landmark constraints (purple edges) and soft landmark-landmark constraints (red edges) between poses as well as between landmarks.
The pose graph optimization occurs using loop closure (light brown edges).}
\vspace{-3mm}
\label{fig:flowchart}
\end{figure*}

Our proposed SoMaSLAM builds on the previous efficient graph SLAM (EG-SLAM)~\cite{zhou2022efficient}.
The shortage of data negatively impacts the result.
We incorporate a soft MW assumption through landmark-landmark constraints.
Sec.~\ref{subsec:efficient2dgraphslam} gives a short explanation regarding \cite{zhou2022efficient}, focusing on sections relevant to our method.
Sec.~\ref{subsec:landmarklandmarkconstraints} gives a thorough explanation of our proposed method.
Fig.~\ref{fig:flowchart} shows a brief overview of the proposed SoMaSLAM method.

\subsection{Efficient 2D Graph SLAM for Sparse Sensing}
\label{subsec:efficient2dgraphslam}

We summarize EG-SLAM briefly (for full details, refer to~\cite{zhou2022efficient}).
EG-SLAM builds upon a graph-based framework with key improvements both in its front-end and back-end.
In the front-end, the method accumulates sparse range data over multiple poses to form a multiscan~\cite{beevers2006slam}, later turned into line segments~\cite{borges2000split, fischler1981ransac}. 
Each line is parameterized as polar coordinates $l = (\rho, \theta)$ and is associated with an existing landmark in the graph.
If there is no appropriate landmark, the line segment creates a new landmark.
This step accounts for line 1, and 2 in Algorithm 1.
A pose and a landmark form a pose-landmark constraint, which is then inserted into the landmark graph.
This step accounts for line 3 in Algorithm 1.
When the pose-landmark constraint is formed, the error term for the constraint is calculated as the following \cite{zhou2022efficient}.
\begin{equation} \label{eq:relative_eq}
e_l(\mathbf{x}_i, \mathbf{l}_j) = \mathbf{v}_{ij} - f(\mathbf{x}_{i}^{-1}, \mathbf{l}_j)    
\end{equation}
where $\mathbf{v}_{ij} = \left( \rho_{ij}, \alpha_{ij} \right)$ is the measurement of the landmark $\mathbf{l}_j$ seen in the frame of $\mathbf{x}_i$. $f(\mathbf{x}_{i}^{-1}, \mathbf{l}_j)$ transforms the current estimate of landmark $\mathbf{l}_j$ from the global frame to the frame of $\mathbf{x}_i$.
In the back-end, the landmark graph provides odometry constraints to the pose graph, also known as loop closing.
The key idea is to simultaneously maintain two graphs, a landmark graph and a pose graph, to optimize both local poses and global loop closures given sparse and noisy data.

\subsection{Landmark-Landmark Constraints on Soft MW}
\label{subsec:landmarklandmarkconstraints}

\begin{algorithm}[t]
\caption{Creating Constraints in Landmark Graph}
\begin{algorithmic}[1]
\For{each segment in segments}
    \State Associate line segments with landmark
    \State Create a pose-landmark constraint and insert into landmark graph
    \For{landmark in landmarks}
        \If{requirements are satisfied}
            \State Create a landmark-landmark constraint and insert into landmark graph
        \EndIf
    \EndFor
\EndFor
\end{algorithmic}
\end{algorithm}

\subsubsection{Landmark-Landmark Constraints}


Measurements between poses and between a pose and a landmark are essential for forming constraints in graph SLAM.
Unlike pose-pose and pose-landmark constraints, constraints between two landmarks cannot be directly formed. Our approach uses the relationship between landmarks based on the Manhattan world assumption~\cite{coughlan1999manhattan} to calculate ideal values, which are then treated as measurements for forming constraints.

\subsubsection{Soft Manhattan World Assumption}
Many environments suitable for mapping resemble a Manhattan world, where building walls are typically rectangular.
However, real-world settings are seldom identical to this structure.
The soft Manhattan world assumption addresses this by maintaining the general structure of a Manhattan world while allowing for non-Manhattan features like curved or diagonal surfaces, effectively capturing the complexity of real environments such as libraries.


\subsubsection{Going over Created Landmarks (line 4)}
We go over existing landmarks to find a landmark capable of forming constraints with the landmark associated with the line segment in question.
The landmark associated with the line segment is not always the most recently created.

\subsubsection{Requirements for Creating Landmark-Landmark Constraints (line 5)} 
This process determines whether the two landmarks are suitable for a landmark-landmark constraint.
We use three criteria to decide:
\begin{enumerate}
    \item Two landmarks are \textit{near} each other
    \item Both landmarks are \textit{significant} landmarks
    \item The landmark-landmark constraint is not oversaturated
\end{enumerate}
Criterion 1 establishes local landmark-landmark constraints from both spatial and temporal perspectives, unlike existing methods such as linear four-point LiDAR SLAM (FL-SLAM)~\cite{jeong2023linear} that rely on global references to align walls with the Manhattan world~\cite{coughlan1999manhattan}.
Here, $\mathbf{l}_1$ represents an existing landmark, while $\mathbf{l}_2$ refers to the current landmark associated with the line segment.

\begin{figure}[t]
\centering
\includegraphics[width=\linewidth]{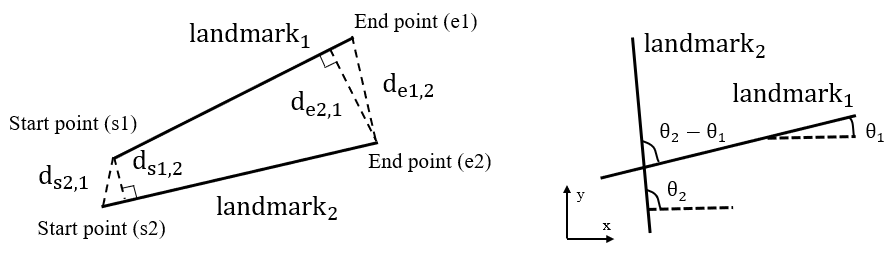}
\vspace{-6mm}
\caption{Visualization of how four distances are measured between two landmarks (left).
$\textup{d}_{m,n}$ is the distance between point $m$ and landmark segment $\mathbf{l_n}$.
Orientation relationship between two landmarks (right) where $\mathbf{\theta_i}$ is the slope angle of i-th landmark $\mathbf{l_i}$.}
\vspace{-3mm}
\label{fig:criterion1}
\end{figure}

We consider landmark segments to be spatially local if they are close to each other or intersect.
For proximity, at least one of the four distances (see Fig.~\ref{fig:criterion1}) must be below a certain threshold.
The landmarks are temporally local if $\mathbf{0} < \mathbf{ID}_{2} - \mathbf{ID}_{1} < n$, where $n \in \mathbb{Z}^+$. 
$\mathbf{ID}_{i}$ represents the creation ID assigned to $\mathbf{l}_i$ in increasing order.
The landmarks must be both temporally and spatially local for criterion 1 to be satisfied.



\begin{figure}[t]
\centering
\includegraphics[width=\linewidth]{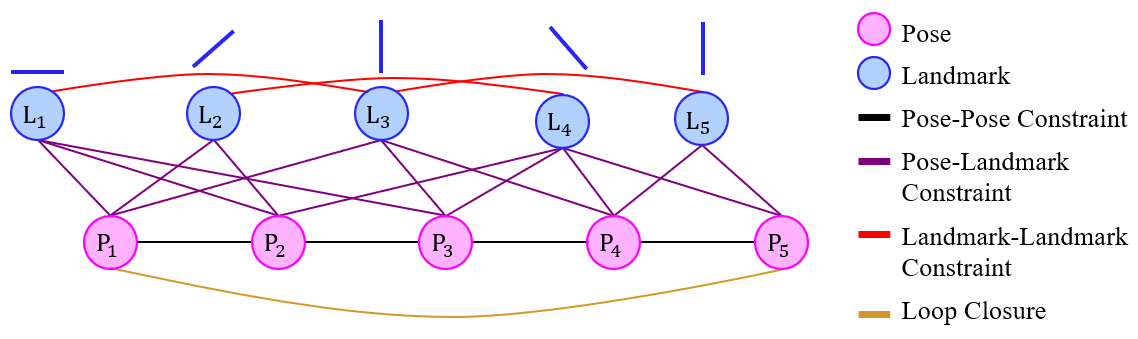}
\vspace{-7mm}
\caption{Pose and landmark graph representation for our SoMaSLAM, with pose-landmark constraints (purple edges), soft landmark-landmark constraints (red edges), and loop closure constraints (light brown edges).
The orientation of the landmarks (blue) and the constraints between them are shown as examples above each landmark node.}
\label{fig:factor_graph_diagram}
\vspace{-5mm}
\end{figure}

Criterion 2 focuses on selecting major landmarks that effectively represent the environment.
To qualify, a landmark must exceed a length threshold and have a sufficient number of pose-landmark constraints connected to it.

Criterion 3 addresses oversaturation by tracking and prevents an excessive number of repeated constraints between the same pair of landmarks, which could cause significant errors during optimization.

\subsubsection{Creating a Landmark-Landmark Constraint (line 6)}

\begin{equation} \label{eq:determine_ideal_angle}
\begin{split}
\left| \mathbf{\theta_2 - \theta_1} - \frac{k}{2}\pi \right| < \epsilon \Rightarrow \mathbf{\Delta\theta_{ideal}} = \frac{k}{2}\pi
\end{split}
\end{equation}


We introduce artificial ideal values to use as measurements to compensate for the absence of actual data.
We focus on the orientation of the landmarks.
The relationship between $\mathbf{l}_1$ and $\mathbf{l}_2$ decides $\mathbf{\Delta\theta_{ideal}}$ in Eq.~\eqref{eq:determine_ideal_angle}, where $\exists k \in \mathbb{Z}$, $\ |k| \leq 2$, and $\epsilon$ is a threshold.
$\mathbf{\Delta\theta_{ideal}}$ is the difference in orientation in an ideal MW where landmarks that are \textit{almost} parallel or orthogonal are treated as \textit{exactly} parallel or orthogonal.

\begin{figure}[t]
\centering
\includegraphics[width=\linewidth]{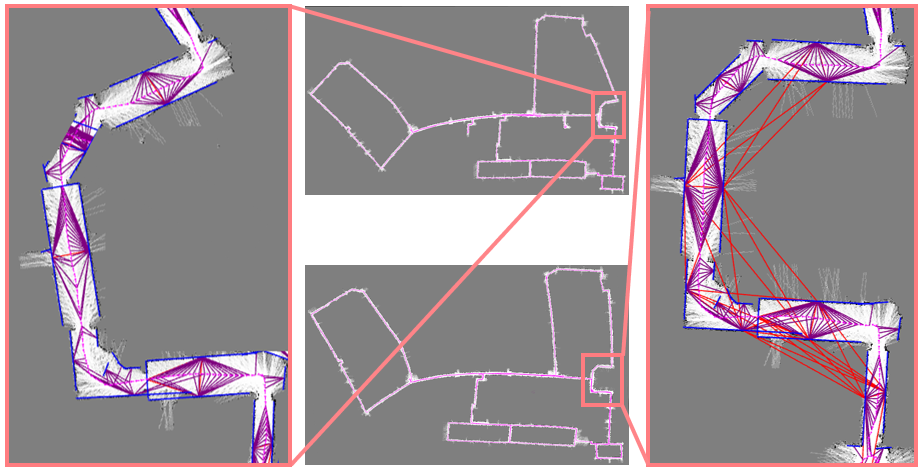}
\vspace{-6mm}
\caption{SoMaSLAM results on MIT Killian with two variables (left) and one variable (right) in the constraint error terms.
Formulating a landmark-landmark constraint with one model parameter shows better performance.}
\vspace{-3mm}
\label{fig:pl_inspiration}
\end{figure}

We draw inspiration from pose-landmark constraints to define the error term for landmark-landmark constraints. 
In pose-landmark constraints, the error term captures the difference between the measured and estimated landmark positions relative to the pose frame.
In our approach, we treat $\mathbf{l}_1$ as the pose and calculate the necessary values for $\mathbf{l}_2$.
Both the measured and estimated orientation of $\mathbf{l}_2$ are relative to $\mathbf{l}_1$.
The ideal orientation of $\mathbf{l}_2$ ($\mathbf{\theta_{2, ideal}}$) is computed as $\mathbf{\theta_{2, ideal}} = \mathbf{\Delta\theta_{ideal}} + \mathbf{\theta_1}$ and is treated as the measurement.
In addition, using one variable in the constraint error term produces a smaller overall error compared to having two variables inside the error term.
Since multiple variables increase degrees of freedom and complexity, they also raise the accumulated error during optimization.
Unnecessary constraint removal during consistency checks~\cite{graham2015robust} can be avoided by reducing the overall error, resulting in a more accurate graph.
Numerical analysis shows that treating both landmarks as variables decreases the number of landmark-landmark constraints by 13.8\% from 4832 to 4166 in the MIT Killian dataset.
It also removes 1\% of existing pose-landmark constraints, negatively impacting the algorithm performance.
Fig.~\ref{fig:pl_inspiration} demonstrates the superior performance of using one variable in the error term compared to using two variables.
\begin{equation} \label{eq:error_for_ll_constraint}
\vspace{-1mm}
e_{ll}(\mathbf{l}_1, \mathbf{l}_2) = \mathbf{\theta_{2, ideal}} - \mathbf{\theta_2}
\vspace{-1mm}
\end{equation} 
We define the error term for landmark-landmark constraints as Eq. \eqref{eq:error_for_ll_constraint} where $e_{ll}$ is the error term for the landmark-landmark constraint, $\mathbf{\theta_{2, ideal}}$ serves as the measured value, and $\mathbf{\theta_2}$ is the estimated value.

\begin{equation} \label{eq:information_matrix}
\mathbf{\Omega_{i,j}} =
\begin{bmatrix}
    \mathbf{len}_1 + \mathbf{len}_2 & 0 \\
    0 & \mathbf{len}_1 + \mathbf{len}_2
\end{bmatrix}
\end{equation}


In graph SLAM, an edge includes both the error term and the information matrix, the latter reflecting the significance of the constraint.
We define our matrix to assign the appropriate importance to each constraint.
Eq.~\eqref{eq:information_matrix} represents the information matrix for a landmark-landmark constraint, where $\mathbf{\Omega_{i,j}}$ is the matrix for a constraint between $\mathbf{l}_i$ and $\mathbf{l}_j$, with $\mathbf{len}_i$ denoting the length of $\mathbf{l}_i$.
Shorter landmarks yield smaller information matrices, while longer landmarks, often representing key features like walls, have larger matrices.
This ensures the optimization process appropriately prioritizes necessary constraints.
Fig.~\ref{fig:factor_graph_diagram} shows the pose graph representation for our proposed SoMaSLAM.


\subsubsection{Graph Optimization}
We use the previously defined $e_{ll}(\mathbf{l}_i, \mathbf{l}_j$) and $\mathbf{\Omega_{i,j}}$ to define the objective function as follows:
\begin{equation} \label{eq:objective_function}
G(X) = \sum_{ij} e_{ll}(\mathbf{l}_i, \mathbf{l}_j)^{T}\mathbf{\Omega_{i,j}} e_{ll}(\mathbf{l}_i, \mathbf{l}_j) + F(X)
\end{equation}
where \textit{F(X)} is the summation of the squared odometry and landmark errors, weighted by their corresponding covariances in EG-SLAM~\cite{zhou2022efficient}.
The Levenberg-Marquardt solver in g2o~\cite{kummerle2011g2o} is used for pose graph optimization.

\subsubsection{Impact of Landmark-Landmark Constraints}

\begin{figure}[t]
\centering
\includegraphics[width=\linewidth]{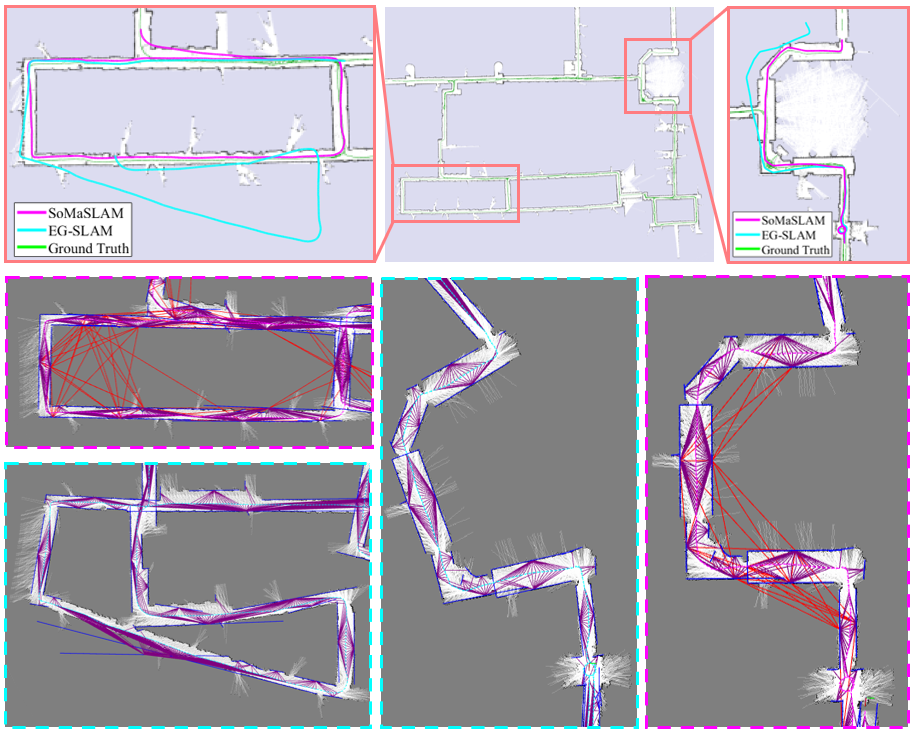}
\vspace{-6mm}
\caption{Estimated poses and landmarks with EG-SLAM (cyan) and SoMaSLAM (magenta) from landmark graphs compared to the ground truth (green) on MIT Killian.
Our proposed SoMaSLAM shows more accurate and consistent mapping results than EG-SLAM thanks to the soft landmark-landmark constraints (red edges).}
\vspace{-3mm}
\label{fig:MIT_Killian_zoom}
\end{figure}



We illustrate the effectiveness of the proposed soft landmark-landmark constraints in Fig.~\ref{fig:MIT_Killian_zoom}.
As noted in Fig.~\ref{fig:flowchart} and Sec.~\ref{subsec:efficient2dgraphslam}, the pose graph performs loop closing based on the relative poses of the landmark graph.
We compare the poses of the landmark graph to demonstrate the effectiveness of landmark-landmark constraints.
These constraints significantly reduce error accumulation and thus yield poses close to that of the ground truth.

\section{Experiments}
\label{sect:experiments}

To demonstrate the effectiveness of the proposed SoMaSLAM, we thoroughly test our algorithm using both Radish \cite{Radish} and author-collected datasets.
We compare SoMaSLAM with EG-SLAM~\cite{zhou2022efficient} and FL-SLAM~\cite{jeong2023linear}.


\subsection{Radish Datasets}



\begin{figure}[t]
\centering
\includegraphics[width=\linewidth]{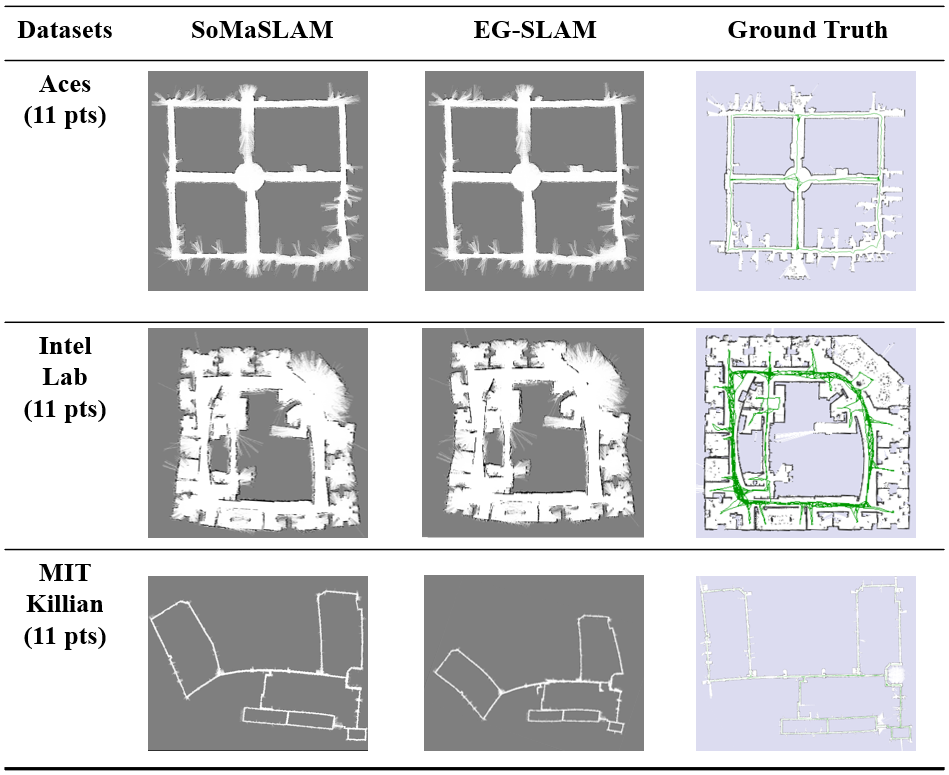}
\vspace{-6mm}
\caption{SLAM results on Radish datasets with SoMaSLAM and EG-SLAM.}
\vspace{-2mm}
\label{fig:radish_11pts}
\end{figure}

Fig.~\ref{fig:radish_11pts} and Table~\ref{tab:radish_11pts} show the qualitative and quantitative results for running SoMaSLAM and EG-SLAM with the ground-truth~\cite{kummerle2009measuring}.
We uniformly sample 11 out of 180 scan points to simulate a sparse range sensing dataset.
Both methods perform well except the MIT Killian dataset.
There is a lack of finesse with the EG-SLAM method, highlighted by the large translational error value, which is 14 times that of SoMaSLAM.
Both methods disappoint when mapping the upper left rectangle area of the MIT Killian dataset, confirmed by a relatively large rotational error.

\begin{table}[t]
\centering
\caption{Evaluation Results on Radish Dataset (11 Pts)}
\begin{tabular}{@{}lcc@{}}
\toprule
                         & \textbf{SoMaSLAM} & \textbf{EG-SLAM} \\ \midrule
\multicolumn{3}{l}{\textbf{Aces}} \\
\quad Absolute translational (m)  & $\mathbf{0.04 \pm 0.05}$    & $0.05 \pm 0.05$    \\
\quad Absolute rotational ($^{\circ}$)    & $\mathbf{1.15 \pm 1.43}$      & $1.16 \pm 1.44$      \\ \midrule

\multicolumn{3}{l}{\textbf{Intel Lab}} \\               
\quad Absolute translational (m)  & $\mathbf{0.09 \pm 0.13}$    & $0.09 \pm 0.13$    \\
\quad Absolute rotational ($^{\circ}$)    & $2.41 \pm 2.50$      & $\mathbf{2.38 \pm 2.54}$      \\ \midrule

\multicolumn{3}{l}{\textbf{MIT Killian}} \\
\quad Absolute translational (m)  & $\mathbf{0.07 \pm 0.15}$    & $0.93 \pm 3.39$    \\
\quad Absolute rotational ($^{\circ}$)    & $\mathbf{1.98 \pm 3.50}$      & $2.29 \pm 4.25$      \\ \bottomrule
\end{tabular}
\vspace{-2mm}
\label{tab:radish_11pts}
\end{table}

\begin{figure}[t]
\centering
\includegraphics[width=\linewidth]{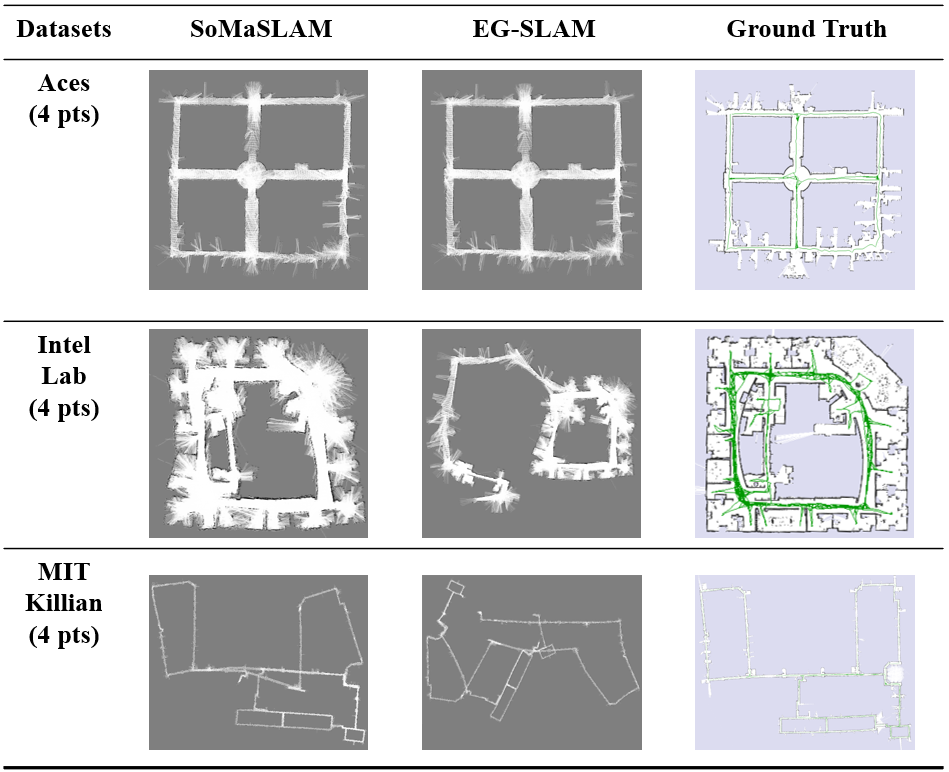}
\vspace{-6mm}
\caption{SLAM results on Radish datasets with SoMaSLAM and EG-SLAM.}
\vspace{-2mm}
\label{fig:radish_4pts}
\end{figure}

\begin{table}[t]
\centering
\caption{Evaluation Results on Radish Dataset (4 Pts)}
\begin{tabular}{@{}lcc@{}}
\toprule
                         & \textbf{SoMaSLAM} & \textbf{EG-SLAM} \\ \midrule
\multicolumn{3}{l}{\textbf{Aces}} \\
\quad Absolute translational (m)  & $0.04 \pm 0.05$    & $\mathbf{0.04 \pm 0.05}$    \\
\quad Absolute rotational ($^{\circ}$)    & $\mathbf{1.04 \pm 1.34}$      & $1.04 \pm 1.36$      \\ \midrule

\multicolumn{3}{l}{\textbf{Intel Lab}} \\               
\quad Absolute translational (m)  & $\mathbf{0.13 \pm 0.21}$    & $2.80 \pm 8.98$    \\
\quad Absolute rotational ($^{\circ}$)    & $\mathbf{2.71 \pm 3.01}$      & $15.19 \pm 36.00$      \\ \midrule

\multicolumn{3}{l}{\textbf{MIT Killian}} \\
\quad Absolute translational (m)  & $\mathbf{0.91 \pm 4.49}$    & $17.84 \pm 46.25$    \\
\quad Absolute rotational ($^{\circ}$)    & $\mathbf{2.19 \pm 3.77}$      & $15.81 \pm 37.69$      \\ \bottomrule
\end{tabular}
\vspace{-2mm}
\label{tab:radish_4pts}
\end{table}

\begin{figure*}[t]
\centering
\includegraphics[width=\linewidth]{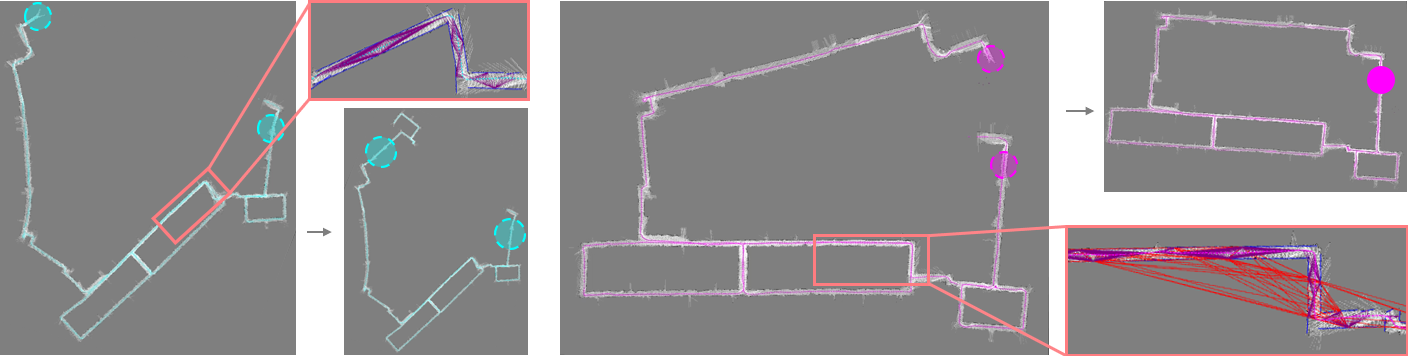}
\vspace{-6mm}
\caption{MIT Killian (4 points) partially mapped with EG-SLAM (left, cyan) and SoMaSLAM (right, magenta).
Loop closure opportunities are translucent circles and successful loop closures are solid circles (magenta).
EG-SLAM (left) fails to add loop closure constraints due to errors accumulating over time, while our SoMaSLAM (right) succeeds thanks to soft landmark-landmark constraints (red edges).}
\vspace{-4mm}
\label{fig:error_reason}
\end{figure*}

Fig.~\ref{fig:radish_4pts} and Table~\ref{tab:radish_4pts} show the qualitative and quantitative results for running SoMaSLAM and EG-SLAM with the ground-truth~\cite{kummerle2009measuring} while only sampling 4 out of the 180 available data points.
For the Aces dataset, SoMaSLAM and EG-SLAM demonstrate strong quantitative and qualitative results.
The performance of SoMaSLAM on the Intel Lab dataset closely matches the ground truth, unlike EG-SLAM.
EG-SLAM exhibits a translational error 20 times greater than SoMaSLAM and a rotational error five times that of SoMaSLAM.
Additionally, the rotational error of EG-SLAM has a striking standard deviation of 36$^{\circ}$.
Both algorithms struggle with the MIT Killian dataset, though SoMaSLAM performs closer to the ground truth while EG-SLAM performs significantly worse.
The translational and rotational errors of EGSLAM are approximately 19 and 7 times greater than those of SoMaSLAM, as shown in Table \ref{tab:radish_4pts}.

The translational and rotational errors from the MIT Killian dataset (4 points) with SoMaSLAM are comparable to the errors from the same dataset (11 points) with EG-SLAM.
This strongly indicates the effectiveness of SoMaSLAM at enhancing SLAM performance with sparse range data.

Fig.~\ref{fig:error_reason} shows the two types of failure cases with one being an extension of the other.
Lack of data and therefore an insufficient number of useful constraints accumulates drift error as shown left of Fig.~\ref{fig:error_reason}.
When this mismanagement continues, the correlative scan-to-map matching~\cite{olson2009real} score decreases.
The score may fall below a certain threshold, which prevents necessary loop closing from occurring shown left of Fig.~\ref{fig:error_reason}.
With the help of landmark-landmark constraints, SoMaSLAM tones down the drift error shown right of Fig.~\ref{fig:error_reason}.
In addition, SoMaSLAM continues to keep the error accumulation in check and successfully performs loop closure, accentuated by a solid magenta circle.


\begin{table}[t]
\centering
\caption{Types of Processing Time per Frame in ms}
\vspace{-2mm}
\begin{tabular}{@{}lccc@{}}
\toprule
                     & \textbf{Aces} & \textbf{Intel Lab} & \textbf{MIT Killian} \\ \midrule
\multicolumn{4}{l}{\textbf{EG-SLAM}} \\
\quad Average & 0.8179 & 2.6253 & 4.7549 \\
\quad Max. Front-End & 20.454 & 17.241 & 98.143 \\
\quad Max. Backend  & 73.823 & 276.36 & 85.262 \\ \midrule

\multicolumn{4}{l}{\textbf{SoMaSLAM}} \\
\quad Average & 1.0668 & 2.7778 & 5.5285 \\
\quad Max. Frontend & 20.454 & 15.181 & 63.079 \\
\quad Max. Backend  & 113.42 & 275.69 & 369.67 \\ 
\bottomrule
\end{tabular}
\label{tab:slam_times}
\end{table}


Table \ref{tab:slam_times} compares the mean processing time per frame for EG-SLAM and SoMaSLAM on an Intel Core i5-12400F PC.
A frame is a single cycle of sensor data processed in the SLAM system. 
The average time difference is negligible.
SoMaSLAM has a higher maximum backend processing time for larger datasets like Intel Lab and MIT Killian, while EG-SLAM shows a significantly higher maximum frontend processing time—up to 55.6\% for MIT Killian and 13.6\% for Intel Lab compared to SoMaSLAM.

\subsection{Author-Collected Datasets}

\begin{figure}[t]
\centering
\includegraphics[width=\linewidth]{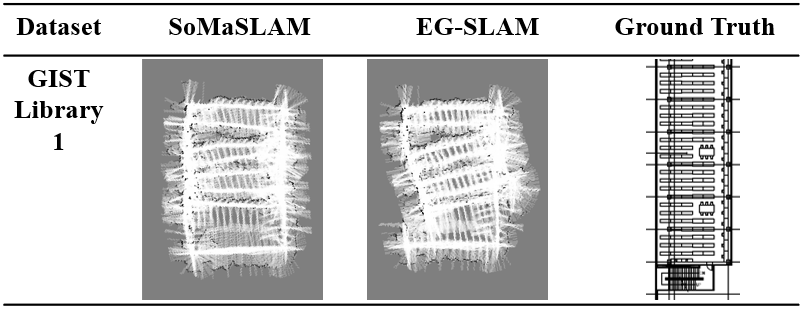}
\vspace{-6mm}
\caption{GIST Library 1 dataset run using SoMaSLAM and EG-SLAM.
The ground truth (right) is a floor plan.}
\label{fig: LG_Library_1}
\vspace{-3mm}
\end{figure}

\begin{figure}[t]
\centering
\includegraphics[width=\linewidth]{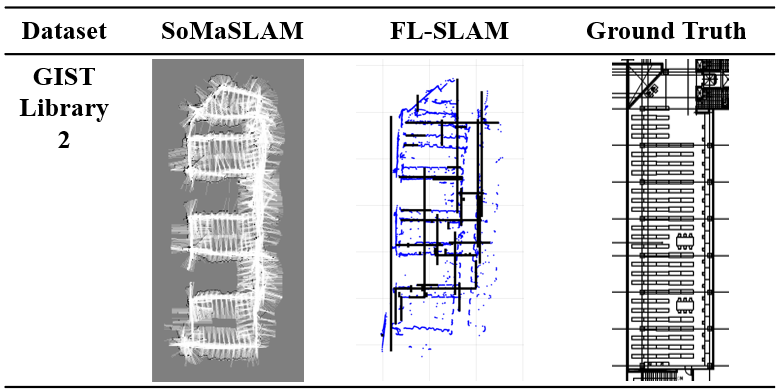}
\vspace{-5mm}
\caption{GIST Library 2 dataset run using SoMaSLAM and FL-SLAM (middle).
FL-SLAM plots the pointcloud (blue) and creates walls (black).}
\vspace{-3mm}
\label{fig: LG_Library_2}
\end{figure}

We employ the Crazyflie nano drone~\cite{giernacki2017crazyflie} to collect sparse range data from the 2nd floor of a library at the Gwangju Institute of Science and Technology (GIST).
We attach an optical flow sensor, a flow deck (1.6 g) for odometry data, and a multi-ranger deck (2.3 g) to collect sparse range measurements from the front, back, left, and right of the Crazyflie.
Since they are low-cost, lightweight, and compact-sized sensors, the range data is very sparse and not accurate enough.
We gather two notable datasets, GIST Library 1 and GIST Library 2, to compare with EG-SLAM and FL-SLAM, respectively.
The sparse range data from the Crazyflie provides the ultimate challenge, serving as a true test to see if the algorithm is viable with real-world sparse data.



EG-SLAM falls short in mapping a structured world shown in Fig.~\ref{fig: LG_Library_1}.
However, SoMaSLAM leverages the distinct Manhattan world-like environment to clearly demonstrate the effectiveness of SoMaSLAM.

FL-SLAM has difficulty mapping sparse, inaccurate Crazyflie data, as shown in Fig.~\ref{fig: LG_Library_2}.
FL-SLAM fails catastrophically from afar.
Upon close inspection, FL-SLAM detects some walls in the MW-like areas of GIST Library 2.
However, it fails to detect the non-MW features inherently at the top and instead creates walls that align with the initially declared Manhattan world.
In contrast, SoMaSLAM successfully maps the environment.

\section{Conclusion}
\label{sect:conclusion}

We introduce two novel ideas in graph SLAM: a soft landmark-landmark constraint and a soft Manhattan world (MW) assumption.
We show how our new method improves the accuracy of graph SLAM with sparse sensing and provides more versatility compared to using strict structural regularities.
In future works, we hope to use multiple drones to effectively address the sparse sensing problem.

\section*{Acknowledgements}
The authors thank the library staff at the GIST Library Yeon Sook Ryu, and Yong Kwan Kim for kindly allowing us to collect data using a Crazyflie drone in the library. We also thank Eunju Jeong for providing the code for \cite{jeong2023linear}.

\bibliographystyle{IEEEtran}
\bibliography{jeahnhan_icra2025}

\end{document}